\documentclass{article} 
\usepackage{iclr2027_conference_new,times}

\usepackage{amsmath,amsfonts,bm}

\def\eqref#1{equation~\ref{#1}}

\def\1{\bm{1}}

\DeclareMathAlphabet{\mathsfit}{\encodingdefault}{\sfdefault}{m}{sl}
\SetMathAlphabet{\mathsfit}{bold}{\encodingdefault}{\sfdefault}{bx}{n}

\usepackage{hyperref}
\usepackage{url}

\usepackage{graphicx}
\usepackage{amsthm}
\theoremstyle{plain}

\theoremstyle{definition}

\usepackage{wrapfig}
\usepackage{algorithm}
\usepackage{algorithmic}
\usepackage{tabularray}
\usepackage{amssymb}
\usepackage{subcaption}
\usepackage{multirow}
\usepackage{arydshln}
\usepackage{tabularx}
\usepackage[utf8]{inputenc} 
\usepackage[T1]{fontenc}    
\usepackage{booktabs}       
\usepackage{amsfonts}       
\usepackage{nicefrac}       
\usepackage{microtype}      
\usepackage{xcolor}         
\usepackage{arydshln}
\usepackage{tikz}
\usepackage{pgfplots}

\title{Budget-Aware Predict-and-Verify Theorem Proving with Heterogeneous LLM Solvers}

\author{Wei Jiang\\
Futurewei Technologies\\
San Jose, CA 95131, USA \\
\texttt{wjiang@futurewei.com} \\
\And
Ziran Liu\\
Shanghai Institute for Mathematics and Interdisciplinary Sciences\\
Shanghai, China\\
\texttt{zliu@simis.cn}\\
\And
Wei Wang \\
Futurewei Technologies \\
San Jose, CA 95131, USA \\
\texttt{rickweiwang@futurewei.com} 
}

\iclrfinalcopy 
\begin{document}

\maketitle

\begin{abstract}
Large language model (LLM) theorem solvers differ in both proving capability and inference cost. Applying an expensive solver to every subgoal wastes computation when cheaper models are sufficient, while always starting from the cheapest solver wastes attempts on subgoals that require stronger models. We present BudgetPV, an agentic budget-aware predict-and-verify system that allocates computation across unresolved subgoals and heterogeneous LLM solvers. BudgetPV uses a cross-solver predictor to guide solver selection for each subgoal and revises the allocation as verification feedback becomes available.

The allocation problem changes as proving progresses. Early decisions mainly determine which solver should handle an untried subgoal, while later decisions determine which previously attempted subgoals should receive additional computation. We train the predictor with two objectives for these two stages. Class-weighted cross-entropy improves assignment to less frequent stronger solvers during the initial stage, while plain cross-entropy provides better calibrated probabilities for later allocation among previously attempted subgoals.

We evaluate BudgetPV on Lean Workbook, miniF2F, ProofNet, and PutnamBench using three progressively expanded solver settings: a strong but expensive solver alone, adding a medium-cost solver with intermediate solving capability, and further adding a cheap solver with lower solving capability. BudgetPV requires less LLM generation computation than uniform allocation to solve the same number of subgoals across the evaluated settings. The experiments further show the computation savings obtained as medium- and lower-cost solvers are added.
\end{abstract}

\section{Introduction}
\label{sec:introduction}

Large language models (LLMs) have improved formal theorem proving. Recent systems generate complete proofs, search over proof states, retrieve useful lemmas, learn from proof-assistant feedback, and decompose difficult proofs into smaller steps~\cite{polu2020,yang2023leandojo,xin2024deepseek,xin2024v15,ren2025deepseekv2,lin2025goedel,lin2026goedelv2}. Many of these gains rely on increased computation at inference time. A common approach generates multiple candidate proofs for the same theorem and submits them to a proof assistant, such as Lean, which mechanically checks formal proofs~\cite{demoura2021lean4}. Increasing the number of candidates often raises the probability of finding a valid proof~\cite{brown2024,snell2024}, but also increases inference and verification costs.

Proof decomposition represents a theorem as a set of smaller subgoals that can be solved separately. The Draft, Sketch, and Prove (DSP) method constructs a formal proof sketch, a partial proof that specifies intermediate subgoals, and proves these subgoals independently~\cite{jiang2023dsp}. Recursive and hierarchical proving methods further divide difficult goals into smaller subgoals~\cite{wang2024recursive,dong2024hierarchical,ren2025deepseekv2}. This decomposition changes the unit of computation from the whole proof to individual subgoals.

However, the difficulty of subgoals can vary widely. Some decomposed subgoals can be solved without LLM generation, while others remain unsolved after repeated proof attempts. A theorem is complete only when every required subgoal is proved. Once a subgoal is solved, further computation on it has no value, while repeated proof generation for a subgoal that continues to fail can consume computation that could be used elsewhere. Different subgoals should receive different amounts of computation as the proof progresses.

The allocation problem becomes more complex when multiple LLM solvers with different capabilities and inference costs are available. A stronger solver can solve subgoals that weaker models cannot, but using it for every subgoal wastes computation when a cheaper solver is sufficient.
Conversely, always starting from the cheapest solver wastes attempts on subgoals that require a stronger model. Efficient allocation must decide both which unresolved subgoal should receive computation and which solver should be used.

We propose BudgetPV, an agentic budget-aware predict-and-verify system that manages the available
computation budget across the subgoals of a proof sketch. Each subgoal is first processed using
standard proof procedures provided by a proof assistant, such as Lean. If it remains unsolved, LLM
solvers generate candidate proofs for the proof assistant to check. Rather than assigning the same number of generations to every unresolved subgoal or using the same solver for all subgoals, BudgetPV uses a cross-solver predictor to guide the allocation of proof attempts across subgoals and solvers according to predicted success and generation cost. These estimates are updated as generated proofs succeed or fail, so the allocation changes as proving progresses.

The allocation problem also changes during the proving process. Early in the process, most unresolved
subgoals have not yet been attempted, and the main decision is which solver should handle each subgoal. Later, many unresolved subgoals have already received proof attempts, and the decision shifts toward determining which of them should receive additional computation. We train the same
cross-solver predictor with two objectives for these two stages. Class-weighted cross-entropy gives
greater weight to the less frequent stronger-solver classes and improves initial solver assignment,
while plain cross-entropy provides better calibrated probabilities for later allocation among
previously attempted subgoals.

Adaptive budget allocation across subgoals has also appeared in recent theorem solvers. AlphaProof allocates computation within its AND--OR tree search~\cite{hubert2026alphaproof}, while BFS-Prover-V2 allocates solver effort within its planner-based search process~\cite{xin2025bfsproverv2}. In these systems, budget allocation is part of the proving algorithm itself.  BudgetPV instead manages budget allocation on top of the underlying proving process. It observes the current subgoals and proof-attempt history, uses its cross-solver predictor to evaluate available subgoal--solver actions, and decides where additional computation should be spent. BudgetPV acts as an agent over the proving process, repeatedly selecting a subgoal--solver action, observing its verification result, and using the updated state to make the next allocation decision.

We evaluate BudgetPV on miniF2F~\cite{zheng2022},
ProofNet~\cite{azerbayev2023}, PutnamBench~\cite{tsoukalas2024}, and Lean Workbook~\cite{ying2024}, 
using three settings that progressively introduce solvers with different capability and cost:
a strong but expensive solver alone, adding a medium-cost solver with intermediate solving capability,
and further adding a cheap solver with lower solving capability. These solvers are
Goedel-Prover-V2-32B~\cite{lin2026goedelv2}, Goedel-Prover-V2-8B~\cite{lin2026goedelv2}, and Kimina-Prover-Distill-8B \cite{kimina2025prover}. BudgetPV is tested in both live and replay settings. The live closed-loop evaluation reflects the real proving scenario, where proof candidates are generated during execution, checked by Lean, and the allocation is updated from the observed outcomes. Controlled replay uses the same recorded proof attempts for all compared methods, removing variation from independent generation. Across the evaluated dataset--solver settings, BudgetPV requires less LLM generation computation than uniform allocation to solve the same number of subgoals. The experiments also show the progressive benefit of adding the medium- and lower-cost solvers.

Our contributions are as follows.

\begin{enumerate}

\item We introduce BudgetPV, an agentic system for managing computation jointly across the unresolved subgoals of a proof sketch and LLM solvers with different solving capability and inference cost. BudgetPV works on top of the underlying proving process and uses information produced during proving to guide budget allocation.

\item BudgetPV uses a cross-solver predictor to guide the solver assignment for each unresolved subgoal. We distinguish the initial solver-assignment stage from the later repeated-attempt stage and train the predictor with class-weighted and plain cross-entropy objectives for these two stages, respectively. Verification outcomes update the information used by the agent for subsequent allocation.

\item We evaluate BudgetPV on four theorem-proving datasets using three progressively expanded solver settings: a strong but expensive solver alone, adding a medium-cost solver with intermediate capability, and further adding a cheap solver with lower capability. BudgetPV requires less LLM generation computation than uniform allocation to solve the same number of subgoals, and the experiments show the computation savings obtained by adding medium- and lower-cost solvers.
\end{enumerate}

\section{Related Work}

\paragraph{Neural theorem proving and proof search.}
LLM-based theorem provers generate complete proofs, predict proof steps, or search over intermediate proof states~\cite{polu2020,xin2024deepseek,xin2024v15,lin2025goedel,lin2026goedelv2}. HyperTree Proof Search uses model predictions to guide tree search~\cite{lample2022}, while LeanDojo combines proof generation with retrieval of relevant premises from the Lean library~\cite{yang2023leandojo}. More recent systems also control how computation is distributed during proof search. AlphaProof selects among unresolved subgoals within an AND--OR tree and directs search toward difficult branches~\cite{hubert2026alphaproof}, while BFS-Prover-V2 combines a high-level planner with multiple prover agents to search over planned subgoals~\cite{xin2025bfsproverv2}. In these systems, computation management is part of the proof-search method. BudgetPV instead provides a separate agentic mechanism for managing computation across subgoals and can be used without changing the internal search algorithm of the prover.

\paragraph{Proof decomposition.}
A theorem is usually decomposed into smaller subgoals to handle long and difficult proofs. Draft, Sketch, and Prove (DSP) converts an informal proof into a formal proof sketch whose open subgoals are proved separately~\cite{jiang2023dsp}. POETRY recursively constructs proof sketches and proves intermediate conjectures at successive levels~\cite{wang2024recursive}, while hierarchical proving methods train models to introduce and prove useful intermediate lemmas~\cite{dong2024hierarchical}. DeepSeek-Prover-V2 also uses recursive subgoal decomposition to construct its cold-start training data~\cite{ren2025deepseekv2}. BudgetPV starts from the decomposed subgoals and allocates computation across both unresolved subgoals and available LLM solvers.

\paragraph{Adaptive computation and automated proof procedures.}
A proof assistant can often solve some goals directly using standard automated procedures, without LLM generation. Hammers and related automation have been used in formal theorem proving~\cite{paulson2010}, and THOR combines language models with automated reasoning tools~\cite{jiang2022thor}. More broadly, recent work shows that allocating inference according to problem difficulty can be more efficient than applying the same amount of computation to every input~\cite{snell2024,brown2024}. Related work on input-adaptive computation allocates repeated generations according to predicted per-input success and marginal gain~\cite{damani2025adaptive}.
BudgetPV considers the same general problem in decomposed theorem proving, where computation must be allocated jointly across subgoals and solvers with different capabilities and costs.

\paragraph{Predicting proof success and solver selection.}
Prediction of model correctness or success has been studied using model confidence~\cite{kadavath2022}, calibrated probability estimation~\cite{guo2017}, learned correctness models~\cite{xiao2025generalized}, hidden-state and activation-based predictors~\cite{varshney2026llmrouter,dicicco2026code}, and kNN-based model routing~\cite{li2025rethinking}. These methods estimate whether a model is likely to succeed on a given input and can be used to guide inference-time computation. In formal theorem proving, each generated candidate is checked by the proof assistant, providing direct feedback from previous attempts. BudgetPV uses a cross-solver predictor rather than independent per-solver predictors. It estimates which solver is appropriate for each unresolved subgoal and updates the allocation as verification feedback becomes available.

\begin{figure*}[t]
    \centering
    \includegraphics[width=0.9\textwidth]{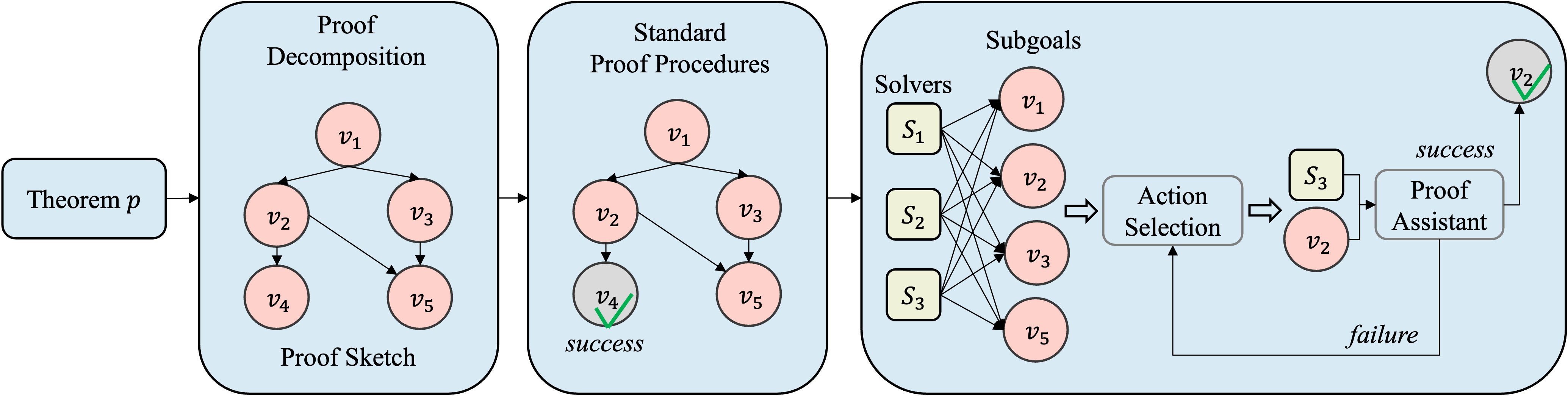}\vspace{-.5em}
  \caption{The overall framework of BudgetPV. A theorem $p$ is decomposed into a proof sketch, and standard proof-assistant procedures first process all subgoals. At each decision step, BudgetPV predicts the success probability of each available LLM solver for each unresolved subgoal and selects a subgoal--solver action based on predicted success, computation cost, and the remaining allowed proof attempts. The selected subgoal is processed using the corresponding solver and verified by the proof assistant. The outcome is used to update the predictions for the next allocation decision.}
  \label{fig:overview}\vspace{-.5em}
\end{figure*}

\section{Method}
\label{sec:method}

BudgetPV is an agentic budget manager for formal theorem proving over decomposed proofs. Given a proof sketch, each subgoal is first processed using standard proof procedures provided by the proof assistant. BudgetPV then manages the computation spent on the unresolved subgoals using the
available LLM solvers.

Figure~\ref{fig:overview} gives an overview of the framework. At each decision step, BudgetPV predicts the probability that each available LLM solver can successfully solve each unresolved subgoal. It then scores the available subgoal--solver actions according to predicted success and generation cost. The selected actions are executed and the generated proofs are checked by the proof assistant. Failed attempts provide additional evidence for the corresponding subgoals, and BudgetPV updates their success estimates before the next allocation decision. This predict-and-verify process continues until the proof is completed or the available computation budget is exhausted.

\subsection{Subgoal--Solver Action Space}
\label{sec:proving_actions}


Given a theorem \(p\), a proof-decomposition method first produces a proof sketch
\(
G_p=(V_p,E_p),
\)
where each node \(v_i\in V_p\) is a formal subgoal and an edge
\((v_i,v_j)\in E_p\) indicates that the result of \(v_i\) is used in proving
\(v_j\). Existing decomposition methods, such as DSP, recursive proving, or hierarchical proving~\cite{jiang2023dsp,wang2024recursive,dong2024hierarchical}, can be used to generate the proof sketch. For each subgoal, BudgetPV records its current status and the results of previous proof attempts. A theorem is completed when all required subgoals in its proof sketch have been proved.

BudgetPV first applies standard proof procedures provided by the proof
assistant to every extracted subgoal. Procedures such as \textit{simp}, \textit{aesop}, \textit{omega}, and \textit{nlinarith} can directly solve some subgoals without LLM inference. Subgoals solved at this stage are removed from further processing, and BudgetPV allocates the LLM computation budget only to the remaining unresolved subgoals.

For these unresolved subgoals, BudgetPV can choose among multiple LLM-based solvers. Let
\(\mathcal{S}=\{S_1,\ldots,S_M\}
\) denote the set of \(M\) available LLM solvers. Different solvers may have different proving capabilities, success probabilities, and computation costs for the same subgoal. 

For an unresolved subgoal \(v_i\), selecting solver \(S_j\) defines a \emph{proving action}
\begin{equation}
a_{i,j}=(v_i,S_j).
\nonumber
\end{equation}
A \emph{proof attempt} is one execution of a proving action. Several proof attempts may use the same subgoal--solver pair because an LLM solver can generate multiple candidate proofs for the same subgoal. Each generated proof is independently checked by the proof assistant.

At decision step $t$, let $U_t$ denote the set of remaining unresolved subgoals. BudgetPV considers the candidate proving actions
\[
    \mathcal{A}_t =
    \left\{
    (v_i,S_j)
    \mid
    v_i\in U_t,\;
    S_j\in\mathcal{S},\;
    n_{i,j}^{(t)}<K_{i,j}
    \right\},
\]
where $n_{i,j}^{(t)}$ is the number of attempts already made using solver
$S_j$ on subgoal $v_i$, and $K_{i,j}$ is the maximum number of attempts
allowed for that pair. The same subgoal may have different success
probabilities and costs under different solvers. That is, BudgetPV allocates
computation over subgoal--solver pairs rather than over subgoals alone.

\subsection{Cross-Solver Success Prediction}

BudgetPV uses a single cross-solver predictor to estimate which solvers are suitable for each unresolved subgoal. We index the solver classes $S_1,\ldots,S_M$ in increasing order of generation cost. For subgoal $v_i$ at decision step $t$, let $H_i^{(t)}$ denote the evidence available from previous proof attempts, including their verification results and errors. The predictor $\Theta$
takes $v_i$ and $H_i^{(t)}$ as input and computes a probability distribution over the $M$ solvers and one additional \texttt{none} class:
\[
q^{(t)}_{i,c}=\Theta(c\mid v_i,H_i^{(t)}),
\qquad
c\in\{S_1,\ldots,S_M,\mathrm{none}\}.
\]
For training, the label is the least expensive available solver that closes the subgoal in the recorded proof-attempt pools. The \texttt{none} label is used when none of the available solvers closes the subgoal. Thus, \texttt{none} is an additional prediction class, not a solver.

The class probabilities are converted into a success estimate for each solver. Probability assigned to a subgoal being solvable by a less expensive solver also contributes to the success estimate of each more expensive solver. Let $\widehat{p}_{i,j}^{(t)}$ denote the estimated probability that solver $S_j$ closes subgoal $v_i$ at decision step $t$. Then
\begin{equation}
\widehat{p}_{i,j}^{(t)}
=
\sum\nolimits_{k=1}^{j} q^{(t)}_{i,S_k}.
\label{eq:solverprobability}
\end{equation}

\subsection{Training for Different Stages of Proving}

The allocation decision changes as proving progresses. Early in the process,
most unresolved subgoals have not yet been attempted, so the main decision is
which solver should be used for each subgoal. Later, many unresolved subgoals
have already received proof attempts, and the main decision is which of these
subgoals should receive additional computation. We train two instances of the
same cross-solver predictor architecture for these two stages.

The predictors are trained using only the formal subgoal, without previous proof attempts or proof-assistant feedback. Each training subgoal $v_i$ is assigned a training label $y_i$ that is the least expensive solver that closes $v_i$ in the recorded proof-attempt pools. If none of the available solvers closes $v_i$, then $y_i=\texttt{none}$.

For the initial stage, we use class-weighted cross-entropy. Let $q^{A}_{i,c}=\Theta_A(c\mid v_i)$ denote the probability assigned to class $c$ by the class-weighted predictor. The loss is
\[
\mathcal{L}_{A}
=
-\sum\nolimits_{i=1}^{N}
w_{y_i}\log q^{A}_{i,y_i},
\qquad
w_c=\frac{N}{(M+1)n_c},
\]
where $N$ is the total number of training subgoals and $n_c$ is the number
of training subgoals in class $c$. The weighting gives greater relative
weight to the less frequent solver classes and improves initial solver
assignment.

For the later stage, we use plain cross-entropy. Let $q^{B}_{i,c}=\Theta_B(c\mid v_i)$ denote the probability assigned to class $c$ by the plain-cross-entropy predictor. The loss is
\[
\mathcal{L}_{B}
=
-\sum\nolimits_{i=1}^{N}
\log q^{B}_{i,y_i}.
\]
Plain cross-entropy retains the empirical class distribution and provides better calibrated probabilities for comparing previously attempted subgoals.

At inference time, BudgetPV uses $\Theta_A$ during the initial stage and switches to $\Theta_B$
once the unresolved set becomes dominated by previously attempted subgoals. The complete proving state is carried across the switch, including previous proof attempts, verification results, attempt counts, and computation already spent.

\subsection{Updating Success After Failed Attempts}

A failed proof attempt provides additional information about which solvers are likely to close the subgoal. BudgetPV uses the training proof-attempt pools to estimate how the success probability of each solver changes after successive failed attempts.

Let $h_{j,c,r}$ denote the measured success rate of the $r$-th attempt by solver $S_j$ for training subgoals with class $c$. After the $r$-th attempt by $S_j$ fails on the subgoal $v_i$, the class probabilities are updated according to
\begin{equation}
\widetilde q_{i,c}^{(t+1)}
=
q_{i,c}^{(t)}
\left(1-h_{j,c,r}\right),
\quad q_{i,c}^{(t+1)}
=
\frac{\widetilde q_{i,c}^{(t+1)}}
{\sum_{c'}\widetilde q_{i,c'}^{(t+1)}}.
\label{eq:successupdate}
\end{equation}
A failure reduces the probability of classes under which that solver was expected to succeed. The updated class probabilities are then converted into new per-solver success estimates for the next allocation decision.

\subsection{Budget-Aware Solver Allocation}

For each available subgoal--solver pair $(v_i,S_j)$, BudgetPV uses the predicted success probability together with the generation cost of the solver. Let $d_j$ denote the generation cost of solver $S_j$ (e.g. measured by GPU-seconds). We compute a score for $(v_i,S_j)$
\[
g^{(t)}(i,S_j)
=
{\widehat p_{i,j}^{(t)}}/{d_j}.
\]
BudgetPV ranks the available subgoal--solver pairs by $g^{(t)}(i,S_j)$. A pair is ranked higher when it has a higher predicted probability of success or a lower generation cost. The highest-ranked pairs are selected for the next proof attempts, with at most one solver assigned to each subgoal at a
decision step.

After an attempt is executed, the generated proof is checked by the proof assistant. If it succeeds, the subgoal is removed from the remaining subgoal set $U_t$. If it fails, the attempt and its proof-assistant feedback are added to the evidence state. BudgetPV then updates the class probabilities using Eq.~\ref{eq:successupdate} and computes new per-solver success estimates using Eq.~\ref{eq:solverprobability} for the next decision step $t+1$. BudgetPV repeatedly selects subgoal--solver pairs, verifies the generated proofs, and updates the proving state until all required subgoals are proved or the computation budget is exhausted.

\section{Experiments}
\label{sec:experiments}

\paragraph{Benchmarks.}
We evaluate BudgetPV on miniF2F~\cite{zheng2022}, ProofNet~\cite{azerbayev2023}, PutnamBench~\cite{tsoukalas2024}, and Lean Workbook~\cite{ying2024}. Standard proof procedures provided by the proof assistant are first applied to every subgoal, and the LLM solvers are evaluated on the unresolved subgoals. Lean Workbook is also used to train the cross-solver predictors. The final Lean Workbook proof-attempt pool contains 4,358 unresolved subgoals. We split the data at the problem level into 3,488 training subgoals and 870 evaluation subgoals, so that subgoals from the same problem do not appear in both sets. For each Lean Workbook training subgoal and each solver, we generate eight proof attempts and check every generated proof with the proof assistant. These recorded attempts are used to determine the training label for the cross-solver predictor.

\paragraph{Solvers.}
BudgetPV is tested with three LLM solvers of different proving capabilities and generation costs: Goedel-Prover-V2-32B \cite{lin2026goedelv2}, Goedel-Prover-V2-8B  \cite{lin2026goedelv2}, and
Kimina-Prover-Distill-8B \cite{kimina2025prover}. We consider three progressive solver settings. The first uses Goedel-Prover-V2-32B alone. The second adds Goedel-Prover-V2-8B, and the third further adds Kimina-Prover-Distill-8B. These settings evaluate how BudgetPV uses additional medium- and lower-cost solvers when they become available.

\paragraph{Predictor training.}
Both predictors use Goedel-Prover-V2-8B as the backbone and are trained with LoRA using rank 32 and scaling factor 64. LoRA is applied to the $q$, $k$, $v$, $o$, gate, up, and down projections. We train for three epochs with learning rate $10^{-5}$, batch size 16, and maximum sequence length
1,536. Predictors $\Theta_A$ and $\Theta_B$ use the same training set. During evaluation, BudgetPV uses $\Theta_A$ initially and switches to $\Theta_B$ when the number of solved
subgoals reaches 75\% of the coverage obtained by Goedel-Prover-V2-32B. The complete proving state is preserved across the switch.

\paragraph{Compared methods.}
We compare BudgetPV with \emph{Uniform}, which uniformly allocates proof attempts across the available solvers and unresolved subgoals without using the success predictor. When multiple solvers are available, Uniform rotates among them according to the number of previous attempts, using the lower-cost solver first when the attempt counts are equal.

We additionally report a \emph{Statistics Oracle} reference, which uses the measured success rate of each solver in the complete proof-attempt pools in place of the predicted success probabilities. The oracle reference indicates how computation could be further saved with more accurate success estimation.

\paragraph{Evaluation metric.}
The main computation metric is the GPU-seconds of LLM proof generation required to reach fixed subgoal-solving targets. For each dataset, we use the subgoals closed by Goedel-Prover-V2-32B alone as the reference set. We report the accumulated generation GPU-seconds required by each method to solve 50\%, 60\%, 70\%, 80\%, and 90\% of the subgoals in this reference set.  The success predictor is evaluated using AUROC~\cite{fawcett2006roc} in Appendix~\ref{app:predictor}.

\paragraph{Live execution and controlled replay.}
We evaluate BudgetPV in both live execution and controlled replay. The live closed-loop evaluation reflects the real proving scenario, where the solvers generate proofs that are checked by Lean and the allocation is updated from the observed outcomes. Controlled replay uses recorded proof attempts and their measured costs instead of generating new proofs. This removes differences caused by independent proof generation. Multiple replay runs are constructed by presenting the recorded attempts in different orders, allowing us to measure how performance changes when successful attempts are encountered at different points. Replay also provides the complete outcome pools needed for the oracle comparisons.

\subsection{Live Closed-Loop Evaluation}
\label{sec:live_results}

Table~\ref{tab:live_main} reports the live closed-loop results. Across all four datasets and all three solver settings, BudgetPV requires less generation computation than Uniform at every reported solve rate. From the live results we can see two complementary sources of computation saving. First, BudgetPV already reduces generation cost when only Goedel-Prover-V2-32B is available. Since there is no solver-selection decision in this 1-solver setting, these gains come from allocating proof attempts more effectively across unresolved subgoals. For example, at the 50\% solve rate, BudgetPV reduces generation computation of Uniform from 11,572 to 6,047 GPU-seconds on Lean Workbook, from 4,423 to 1,656 on PutnamBench, from 2,672 to 1,231 on miniF2F, and from 1,086 to 551 on ProofNet.

Second, access to additional lower-cost solvers provides another source of saving. BudgetPV can use these solvers for subgoals that do not require the most expensive model. On Lean Workbook at the 50\% solving-rate target, the computation required by BudgetPV decreases from 6,047 GPU-seconds with one solver to 3,402 with two solvers and 3,213 with three solvers. On PutnamBench, the corresponding values are 1,656, 731, and 654 GPU-seconds. These results show that BudgetPV benefits both from deciding where to spend computation and from deciding which solver should receive that computation.

The savings are largest at lower and intermediate solving-rate targets. At these stages, many unresolved subgoals can still be solved without repeatedly calling the strongest solver. As the solve rate increases, the unresolved set becomes increasingly concentrated on harder subgoals, and additional computation from the stronger solver becomes more important. This explains why the computation difference between BudgetPV and Uniform generally decreases toward the highest solve rates.

\begin{table*}[t]
\centering
\caption{Live closed-loop generation GPU-seconds required to reach the corresponding solving-rate target. Results are mean $\pm$ standard deviation over five runs. Lower is better. }
\label{tab:live_main}
\footnotesize
\setlength{\tabcolsep}{1.6pt}
\begin{tabular}{lclccccc}
\toprule
Dataset & Solvers & Method
& 50\% & 60\% & 70\% & 80\% & 90\% \\
\midrule

miniF2F
& 1 & Uniform
& 2,672$\pm$193 & 3,248$\pm$110 & 3,580$\pm$63
& 3,949$\pm$77 & 5,056$\pm$245 \\
& 1 & BudgetPV
& \textbf{1,231$\pm$70} & \textbf{1,556$\pm$18}
& \textbf{1,955$\pm$139} & \textbf{2,870$\pm$136}
& \textbf{4,881$\pm$649} \\
& 2 & Uniform
& 2,371$\pm$144 & 2,771$\pm$144 & 3,033$\pm$311
& 3,428$\pm$311 & 4,484$\pm$188 \\
& 2 &BudgetPV
& \textbf{1,249$\pm$161} & \textbf{1,390$\pm$83}
& \textbf{1,652$\pm$145} & \textbf{2,282$\pm$144}
& \textbf{4,096$\pm$268} \\
& 3 & Uniform
& 2,267$\pm$106 & 2,446$\pm$210 & 2,812$\pm$109
& 2,989$\pm$200 & 4,195$\pm$269 \\
& 3 &BudgetPV
& \textbf{1,163$\pm$136} & \textbf{1,375$\pm$236}
& \textbf{2,033$\pm$163} & \textbf{2,624$\pm$645}
& \textbf{3,818$\pm$450} \\

\midrule
ProofNet
& 1 & Uniform
& 1,086$\pm$374 & 1,613$\pm$850 & 2,355$\pm$708
& 3,658$\pm$996 & 6,048$\pm$2,305 \\
& 1 &BudgetPV
& \textbf{551$\pm$208} & \textbf{809$\pm$200}
& \textbf{1,063$\pm$111} & \textbf{1,901$\pm$404}
& \textbf{4,203$\pm$705} \\
& 2 & Uniform
& 941$\pm$47 & 1,302$\pm$294 & 2,102$\pm$282
& 3,314$\pm$1,259 & 6,196$\pm$1,262 \\
& 2 & BudgetPV
& \textbf{804$\pm$225} & \textbf{1,032$\pm$71}
& \textbf{1,199$\pm$142} & \textbf{2,466$\pm$680}
& \textbf{3,324$\pm$1,381} \\
& 3 & Uniform
& 1,238$\pm$4 & 1,238$\pm$4 & 2,158$\pm$705
& 4,306$\pm$893 & 8,234$\pm$2,098 \\
& 3 & BudgetPV
& \textbf{711$\pm$195} & \textbf{1,209$\pm$31}
& \textbf{1,955$\pm$898} & \textbf{3,763$\pm$2,802}
& \textbf{6,791$\pm$2,539} \\

\midrule
PutnamBench
& 1 & Uniform
& 4,423$\pm$660 & 6,047$\pm$798 & 7,347$\pm$954
& 14,371$\pm$5,733 & 33,575$\pm$10,280 \\
& 1 & BudgetPV
& \textbf{1,656$\pm$124} & \textbf{4,900$\pm$8}
& \textbf{6,889$\pm$467} & \textbf{12,090$\pm$1,883}
& \textbf{25,604$\pm$6,277} \\
& 2 & Uniform
& 4,353$\pm$750 & 5,865$\pm$636 & 8,765$\pm$2,209
& 14,027$\pm$4,369 & 26,755$\pm$16,452 \\
& 2 & BudgetPV
& \textbf{731$\pm$320} & \textbf{1,311$\pm$459}
& \textbf{5,758$\pm$2,055} & \textbf{11,882$\pm$4,103}
& \textbf{26,684$\pm$5,917} \\
& 3 & Uniform
& 2,769$\pm$379 & 4,698$\pm$1,179 & 8,051$\pm$1,258
& 10,828$\pm$1,511 & 19,677$\pm$1,522 \\
& 3 & BudgetPV
& \textbf{654$\pm$216} & \textbf{1,934$\pm$396}
& \textbf{3,986$\pm$473} & \textbf{8,530$\pm$3,091}
& \textbf{18,329$\pm$828} \\

\midrule
Lean Workbook
& 1 & Uniform
& 11,572$\pm$263 & 13,163$\pm$85 & 15,368$\pm$238
& 17,845$\pm$437 & 32,113$\pm$2,012 \\
& 1 & BudgetPV
& \textbf{6,047$\pm$158} & \textbf{8,268$\pm$370}
& \textbf{10,713$\pm$491} & \textbf{17,140$\pm$478}
& \textbf{31,582$\pm$2,344} \\
& 2 & Uniform
& 8,914$\pm$918 & 10,089$\pm$981 & 11,279$\pm$973
& 14,074$\pm$900 & 22,889$\pm$1,223 \\
& 2 & BudgetPV
& \textbf{3,402$\pm$365} & \textbf{4,526$\pm$121}
& \textbf{6,465$\pm$411} & \textbf{9,585$\pm$414}
& \textbf{19,482$\pm$1,235} \\
& 3 & Uniform
& 7,439$\pm$337 & 8,757$\pm$90 & 11,110$\pm$768
& 15,164$\pm$1,229 & 23,941$\pm$4,703 \\
& 3 & BudgetPV
& \textbf{3,213$\pm$273} & \textbf{4,774$\pm$49}
& \textbf{6,809$\pm$880} & \textbf{10,691$\pm$275}
& \textbf{21,882$\pm$1,428} \\
\bottomrule
\end{tabular}
\end{table*}

\subsection{Controlled Replay and Oracle Evaluation}
\label{sec:replay_results}

Table~\ref{tab:replay_main} reports the controlled replay results together with the oracle reference. The same overall pattern can be observed as the live evaluation. BudgetPV requires less generation computation than Uniform at every reported solving-rate target across all four datasets and
all three solver settings. The reductions are again generally larger at lower and intermediate solving rates. For example, with three solvers, BudgetPV reduces the 50\% computation from 2,275 for Uniform to 997 GPU-seconds on Lean Workbook, from 1,188 to 439 on PutnamBench, from 590 to 517 on ProofNet, and from 341 to 234 on miniF2F. At 90\% solve rate, BudgetPV reduces computation from
16,046 to 13,361 GPU-seconds on Lean Workbook, from 14,439 to 11,002 on PutnamBench, from 5,044 to 2,268 on ProofNet, and from 1,870 to 1,610 on miniF2F.

The agreement between live execution and controlled replay is important because replay removes variation caused by independent proof generation. All methods use the same recorded proof-attempt pools and measured generation costs, while differing only in how computation is allocated.  The consistent advantage of BudgetPV in replay confirms that the savings observed in live execution come from better allocation of proof attempts rather than from random variation in proof generation.

Replay also allows us to compare BudgetPV with the Statistics Oracle, which uses the measured success rates of the solvers in the complete proof-attempt pools. The Statistics Oracle provides a reference for how the allocation could improve if the success estimates were more accurate. Its advantage is most visible at the higher solving-rate targets. With three solvers at the 90\% target, the Statistics Oracle requires 2,878 GPU-seconds on Lean Workbook, compared with 13,361 for BudgetPV. On ProofNet, the corresponding values are 385 and 2,268 GPU-seconds. PutnamBench shows a similar difference, with 669 GPU-seconds for the Statistics Oracle and 11,002 for BudgetPV. The oracle comparison shows that more accurate success estimation could further reduce computation within the same solver setting. BudgetPV already recovers part of the saving by using its learned success estimates, but considerably more computation can be avoided when the success information is more accurate. This suggests that improving cross-solver success estimation is a direct way to further reduce computation without changing the basic predict-and-verify allocation framework.

\begin{table*}[t]
\centering
\caption{Controlled replay generation GPU-seconds required to reach the corresponding solving-rate target. BudgetPV results are mean $\pm$ standard deviation over five runs. Uniform and Statistics Oracle are deterministic for a fixed proof-attempt pool. Lower is better.}\vspace{-.5em}
\label{tab:replay_main}
\footnotesize
\setlength{\tabcolsep}{2pt}
\begin{tabular}{lclrrrrr}
\toprule
Dataset & Solvers & Method
& 50\% & 60\% & 70\% & 80\% & 90\% \\
\midrule

miniF2F
& 1 & Uniform
& 1,243 & 1,385 & 1,558 & 1,724 & 2,756 \\
& 1 & BudgetPV
& \textbf{724$\pm$17} & \textbf{891$\pm$27} & \textbf{1,097$\pm$3} & \textbf{1,605$\pm$2} & \textbf{2,205$\pm$7} \\
& 1 & Statistics Oracle
& 427 & 812 & 954 & 1,118 & 1,257 \\

& 2 & Uniform
& 682 & 743 & 978 & 1,301 & 2,180 \\
& 2 & BudgetPV
& \textbf{334$\pm$1} & \textbf{412$\pm$2} & \textbf{686$\pm$98} & \textbf{929$\pm$3} & \textbf{1,958$\pm$189} \\
& 2 & Statistics Oracle
& 200 & 482 & 626 & 792 & 962 \\

& 3 & Uniform
& 341 & 488 & 700 & 961 & 1,870 \\
& 3 & BudgetPV
& \textbf{234$\pm$1} & \textbf{335$\pm$38} & \textbf{442$\pm$7} & \textbf{741$\pm$83} & \textbf{1,610$\pm$25} \\
& 3 & Statistics Oracle
& 159 & 159 & 450 & 632 & 801 \\

\midrule
ProofNet
& 1 & Uniform
& 1,158 & 1,725 & 2,168 & 3,095 & 4,886 \\
& 1 & BudgetPV
& \textbf{732$\pm$9} & \textbf{1,150$\pm$2} & \textbf{1,508$\pm$14} & \textbf{2,399$\pm$9} & \textbf{3,327$\pm$7} \\
& 1 & Statistics Oracle
& 165 & 274 & 274 & 402 & 448 \\

& 2 & Uniform
& 1,228 & 2,904 & 2,904 & 4,250 & 5,957 \\
& 2 & BudgetPV
& \textbf{929$\pm$9} & \textbf{1,347$\pm$6} & \textbf{1,693$\pm$308} & \textbf{2,490$\pm$231} & \textbf{3,031$\pm$247} \\
& 2 & Statistics Oracle
& 165 & 273 & 273 & 404 & 453 \\

& 3 & Uniform
& 590 & 1,374 & 3,322 & 3,322 & 5,044 \\
& 3 & BudgetPV
& \textbf{517$\pm$34} & \textbf{878$\pm$28} & \textbf{1,218$\pm$283} & \textbf{1,691$\pm$268} & \textbf{2,268$\pm$38} \\
& 3 & Statistics Oracle
& 159 & 159 & 264 & 264 & 385 \\

\midrule
PutnamBench
& 1 & Uniform
& 3,163 & 3,787 & 4,922 & 10,471 & 13,057 \\
& 1 & BudgetPV
& \textbf{845$\pm$0} & \textbf{1,406$\pm$0} & \textbf{4,474$\pm$0} & \textbf{8,009$\pm$619} & \textbf{12,682$\pm$16} \\
& 1 & Statistics Oracle
& 469 & 469 & 469 & 651 & 722 \\

& 2 & Uniform
& 2,320 & 3,253 & 5,376 & 8,418 & 17,062 \\
& 2 & BudgetPV
& \textbf{706$\pm$45} & \textbf{1,498$\pm$427} & \textbf{4,628$\pm$1,674} & \textbf{7,361$\pm$413} & \textbf{10,771$\pm$640} \\
& 2 & Statistics Oracle
& 529 & 529 & 529 & 529 & 712 \\

& 3 & Uniform
& 1,188 & 1,617 & 3,310 & 6,568 & 14,439 \\
& 3 & BudgetPV
& \textbf{439$\pm$81} & \textbf{697$\pm$197} & \textbf{1,831$\pm$142} & \textbf{6,330$\pm$211} & \textbf{11,002$\pm$128} \\
& 3 & Statistics Oracle
& 481 & 481 & 481 & 481 & 669 \\

\midrule
Lean Workbook
& 1 & Uniform
& 5,487 & 6,397 & 7,236 & 8,256 & 14,895 \\
& 1 & BudgetPV
& \textbf{2,445$\pm$207} & \textbf{3,209$\pm$36}& \textbf{4,576$\pm$19} & \textbf{6,468$\pm$4} & \textbf{13,120$\pm$7} \\
& 1 & Statistics Oracle
& 1,915 & 2,940 & 3,477 & 3,857 & 3,996 \\

& 2 & Uniform
& 2,743 & 3,073 & 4,204 & 8,335 & 14,757 \\
& 2 & BudgetPV
& \textbf{1,078$\pm$4} & \textbf{1,654$\pm$33} & \textbf{3,349$\pm$150} & \textbf{6,491$\pm$165} & \textbf{11,796$\pm$141} \\
& 2 & Statistics Oracle
& 1,374 & 2,227 & 2,582 & 2,755 & 3,069 \\

& 3 & Uniform
& 2,275 & 3,584 & 4,418 & 6,805 & 16,046 \\
& 3 & BudgetPV
& \textbf{997$\pm$18} & \textbf{1,717$\pm$98} & \textbf{2,943$\pm$138} & \textbf{5,875$\pm$362} & \textbf{13,361$\pm$384} \\
& 3 & Statistics Oracle
& 904 & 2,004 & 2,399 & 2,576 & 2,878 \\
\bottomrule
\end{tabular}\vspace{-.5em}
\end{table*}

\subsection{Theorem-Level Completion at Matched Generation Budgets}
\label{sec:theorem_level}

Sections~\ref{sec:live_results} and~\ref{sec:replay_results} compare the generation cost of solving individual
subgoals. We also evaluate the number of complete theorems obtained at a
fixed generation budget. The evaluation uses the same proof sketches as
the subgoal experiments. Standard Lean procedures close some subgoals
before LLM generation. A theorem is counted as completed when all of its
remaining subgoals have been closed. The evaluation contains 479 theorems
on Lean Workbook, 151 on miniF2F, 47 on ProofNet, and 200 on PutnamBench.
Of these, 61\%, 56\%, 66\%, and 73\%, respectively, have one open subgoal
after standard Lean procedures. The other theorems require multiple
subgoals to be closed before they count as completed.

For each benchmark and solver configuration, we compare BudgetPV and Uniform at three identical generation budgets. The mean GPU-seconds at which BudgetPV reaches 50\%, 70\%, and 90\% of the number of theorems completed on that benchmark for each solver configuration are used as the budget points. For each budget point, we count completed theorems in the live runs of both methods. Table~\ref{tab:theorem_matched_budget}
reports the mean and standard deviation over three runs.

\begin{table*}[!htbp]
\centering
\caption{Completed theorems under identical generation budgets. Results are mean $\pm$ standard deviation over three live runs.
Higher is better.}
\label{tab:theorem_matched_budget}
\footnotesize
\setlength{\tabcolsep}{0.2pt}
\begin{tabular}{lclccc}
\toprule
Dataset & Solvers & & 50\% & 70\% & 90\% \\
\midrule
miniF2F
& 1 & Budget (GPU-s) & 1,583 & 3,534 & 6,812 \\
& 1 & Uniform
& 20.0$\pm$1.4 & 42.3$\pm$1.2 & 71.3$\pm$0.9 \\
& 1 & BudgetPV
& \textbf{39.7$\pm$2.6} & \textbf{58.3$\pm$4.5}
& \textbf{73.7$\pm$0.9} \\
& 2 & Budget (GPU-s) & 1,652 & 2,945 & 5,894 \\
& 2 & Uniform
& 22.3$\pm$1.2 & 44.0$\pm$3.6
& \textbf{75.7$\pm$2.1} \\
& 2 & BudgetPV
& \textbf{39.7$\pm$1.7} & \textbf{60.7$\pm$3.9}
& 73.0$\pm$2.4 \\
& 3 & Budget (GPU-s) & 1,845 & 2,923 & 4,871 \\
& 3 & Uniform
& 26.7$\pm$0.5 & 56.7$\pm$2.9
& \textbf{74.7$\pm$1.2} \\
& 3 & BudgetPV
& \textbf{40.7$\pm$0.9} & \textbf{60.3$\pm$3.4}
& 73.7$\pm$2.5 \\

\midrule
ProofNet
& 1 & Budget (GPU-s) & 623 & 1,063 & 5,511 \\
& 1 & Uniform
& 2.7$\pm$1.2 & 3.0$\pm$0.8 & 6.7$\pm$0.5 \\
& 1 & BudgetPV
& \textbf{3.3$\pm$0.5} & \textbf{5.7$\pm$0.5}
& \textbf{7.3$\pm$0.5} \\
& 2 & Budget (GPU-s) & 700 & 2,012 & 4,625 \\
& 2 & Uniform
& \textbf{4.0$\pm$0.8} & 5.7$\pm$0.5
& 7.0$\pm$0.8 \\
& 2 & BudgetPV
& 3.7$\pm$0.5 & \textbf{6.3$\pm$1.7}
& \textbf{7.3$\pm$0.5} \\
& 3 & Budget (GPU-s) & 849 & 1,760 & 4,407 \\
& 3 & Uniform
& \textbf{3.3$\pm$0.5} & 5.0$\pm$0.8
& 6.3$\pm$0.5 \\
& 3 & BudgetPV
& 3.0$\pm$2.2 & \textbf{6.0$\pm$0.8}
& \textbf{7.7$\pm$0.5} \\

\midrule
PutnamBench\!\!
& 1 & Budget (GPU-s) & 9,294 & 23,408 & 53,442 \\
& 1 & Uniform
& 3.3$\pm$0.9 & 4.0$\pm$0.8 & \textbf{6.3$\pm$0.9} \\
& 1 & BudgetPV
& \textbf{3.7$\pm$0.5} & \textbf{5.0$\pm$0.8}
& \textbf{6.3$\pm$0.9} \\
& 2 & Budget (GPU-s) & 11,686 & 32,215 & 32,127 \\
& 2 & Uniform
& \textbf{3.3$\pm$0.5} & \textbf{6.0$\pm$1.4}
& \textbf{6.0$\pm$1.4} \\
& 2 & BudgetPV
& 3.0$\pm$0.8 & \textbf{6.0$\pm$1.4}
& \textbf{6.0$\pm$1.4} \\
& 3 & Budget (GPU-s) & 8,381 & 17,813 & 33,091 \\
& 3 & Uniform
& \textbf{4.0$\pm$0.8} & 5.0$\pm$0.0 & 6.0$\pm$0.8 \\
& 3 & BudgetPV
& \textbf{4.0$\pm$0.8} & \textbf{5.7$\pm$1.2}
& \textbf{7.3$\pm$1.2} \\

\midrule
Lean Workbook\!\!\!\!\!\!
& 1 & Budget (GPU-s) & 8,178 & 17,052 & 40,904 \\
& 1 & Uniform
& 50.0$\pm$2.2 & 91.3$\pm$2.5
& 127.3$\pm$1.2 \\
& 1 & BudgetPV
& \textbf{72.0$\pm$2.2} & \textbf{100.3$\pm$1.7}
& \textbf{130.0$\pm$2.2} \\
& 2 & Budget (GPU-s) & 5,208 & 9,575 & 27,840 \\
& 2 & Uniform
& 40.3$\pm$3.1 & 69.3$\pm$6.9
& 125.7$\pm$5.7 \\
& 2 & BudgetPV
& \textbf{72.3$\pm$2.5} & \textbf{100.7$\pm$2.9}
& \textbf{129.7$\pm$3.4} \\
& 3 & Budget (GPU-s) & 5,368 & 10,706 & 39,120 \\
& 3 & Uniform
& 44.0$\pm$5.0 & 80.3$\pm$2.5
& \textbf{135.0$\pm$5.0} \\
& 3 & BudgetPV
& \textbf{71.7$\pm$5.2} & \textbf{101.0$\pm$2.2}
& 130.0$\pm$1.4 \\
\bottomrule
\end{tabular}
\end{table*}

Across the 36 matched-budget comparisons, BudgetPV completes more theorems in 26 cases, matches Uniform in 4 cases, and completes fewer in 6 cases. In the single-solver setting, BudgetPV consistently outperforms Uniform or achieves the same theorem completion count across all 4 benchmarks and
3 budget levels. With additional solvers, BudgetPV generally retains its advantage at the 50\% and 70\% budget levels on Lean Workbook and miniF2F, while the results at higher budgets become less consistent.
On ProofNet and PutnamBench, the improvements are smaller and vary across solver configurations.

With a single solver, BudgetPV only needs to allocate attempts across subgoals according to their predicted success probabilities. Adding heterogeneous solvers introduces the decision of selecting an appropriate solver for each subgoal. Cheaper solvers can reduce generation cost when they can successfully handle the assigned subgoals, but incorrect assignments
consume computation without making progress.
This tradeoff becomes more important as the remaining subgoals become harder and require stronger solvers.

On ProofNet and PutnamBench, Goedel-Prover-V2-32B completes only eight reference theorems on each benchmark, so the completion of even one additional theorem can substantially change the comparison. The smaller number of completed theorems also makes the results more sensitive to individual proof outcomes.

\paragraph{Additional analysis and ablations.}
Appendix~\ref{app:ablations} provides additional experimental details, including the generation settings of evaluated solvers and the experiments used to select the three solvers, as well as the detailed training configuration of the cross-solver predictors.  Appendix~\ref{app:predictor} further evaluates the predictors on the Lean Workbook evaluation set and during live proving. 

\section{Conclusion}
\label{sec:conclusion}

We presented BudgetPV, an agentic predict-and-verify framework for managing computation across unresolved proof subgoals and heterogeneous LLM solvers. BudgetPV estimates the probability that each available solver can close each subgoal and combines this estimate with the measured generation cost of the solver to select subsequent proof attempts. After each attempt is checked by the proof assistant, the resulting evidence is used to update the success estimates and subsequent allocation.

Across four datasets and three progressive solver settings, BudgetPV requires less LLM generation computation than uniform allocation at every reported solving-rate target in both live proving and controlled replay. BudgetPV benefits both from better allocation across unresolved subgoals and from access to lower-cost solvers. With the expensive solver alone, BudgetPV reduces computation by deciding which subgoals should receive additional proof attempts. As medium- and lower-cost solvers are added, it obtains further savings by assigning suitable subgoals to less expensive solvers. The Statistics Oracle further indicates that more accurate success estimation could reduce computation. \vspace{-.5em}

\paragraph{Limitations.}
BudgetPV depends on the accuracy of its success estimates, and the gap to the Statistics Oracle shows that prediction remains an important source of further improvement. The current predictor is trained on Lean Workbook and evaluated on four theorem-proving datasets, so broader training data and
additional solver families may improve its generalization. BudgetPV also operates on a given proof sketch and does not revise the decomposition when repeated failures indicate that a subgoal may be difficult because of the chosen sketch. Future work can jointly consider proof-sketch revision and
computation allocation, improve success estimation from accumulated verification feedback, and evaluate larger solver sets and broader formal reasoning tasks.

\bibliographystyle{plainnat}
\bibliography{ref}

\appendix

\section{Additional Experimental Details and Ablations}
\label{app:ablations}

\subsection{Solver Generation Settings}

Table~\ref{tab:solver_generation_settings} gives the generation settings used for the solvers in our experiments. We follow the recommended generation settings of each model. Goedel-Prover-V2-32B is served with tensor parallelism of two H100 GPUs, while the 8B models use a single H100 GPU for generation.

\begin{table}[t]
\centering
\caption{Generation settings used for the evaluated solvers.}
\label{tab:solver_generation_settings}
\small
\setlength{\tabcolsep}{4pt}
\begin{tabular}{lccc}
\toprule
Solver & Temperature / top-$p$ & Max tokens & Context \\
\midrule
Goedel-Prover-V2-8B
& 1.0 / 0.95 & 40,960 & 40,960 \\
Goedel-Prover-V2-32B
& 1.0 / 0.95 & 40,960 & 40,960 \\
Kimina-Prover-Distill-8B
& 0.6 / 0.95 & 8,096 & 16,384 \\
Pythagoras-Prover-4B
& 1.0 / 0.95 & 30,720 & 32,768 \\
Pythagoras-Prover-32B
& 1.0 / 0.95 & 30,720 & 32,768 \\
DeepSeek-Prover-V2-7B
& 1.0 / 0.95 & 8,192 & 16,384 \\
\bottomrule
\end{tabular}
\end{table}

\begin{table}[t]
\centering
\caption{Solver selection. Close@8 is the percentage of unresolved subgoals
closed within eight proof attempts. Generation cost is the mean GPU-seconds
per attempt measured on Lean Workbook.}
\label{tab:solver_screening}
\small
\setlength{\tabcolsep}{5pt}
\begin{tabular}{lccccc}
\toprule
& \multicolumn{4}{c}{Close@8 (\%)} & \\
\cmidrule(lr){2-5}
Solver
& Lean Workbook
& miniF2F
& ProofNet
& PutnamBench
& GPU-s/attempt \\
\midrule
Pythagoras-Prover-4B
& 31.6 & 67.4 & 18.2 & 11.1 & 3.74 \\
Pythagoras-Prover-32B
& 37.0 & 70.3 & 24.6 & 14.1 & 8.19 \\
Goedel-Prover-V2-8B
& 40.8 & 71.6 & 23.4 & 14.3 & 5.00 \\
Goedel-Prover-V2-32B
& 43.2 & 75.8 & 33.8 & 17.2 & 10.44 \\
DeepSeek-Prover-V2-7B
& 37.2 & 69.0 & 26.0 & 13.3 & 4.21 \\
Kimina-Prover-Distill-8B
& 30.2 & 66.5 & 20.8 & 12.2 & 2.07 \\
\bottomrule
\end{tabular}
\end{table}

\begin{table}[h]
\centering
\caption{Computation saving from adding a lower-cost candidate to
Goedel-Prover-V2-8B. Values report the reduction in GPU-seconds required to
reach the full coverage of Goedel-Prover-V2-8B. Positive values indicate a
saving.}
\label{tab:low_cost_selection}
\small
\setlength{\tabcolsep}{4pt}
\begin{tabular}{lrrrr}
\toprule
Candidate & LWB & miniF2F & ProofNet & PutnamBench \\
\midrule
Kimina-Prover-Distill-8B
& +10\% & +30\% & +28\% & +13\% \\
Pythagoras-Prover-4B
& -1\% & -3\% & -12\% & +1\% \\
DeepSeek-Prover-V2-7B
& -24\% & -20\% & +35\% & -- \\
\bottomrule
\end{tabular}
\end{table}

\subsection{Solver Selection}
\label{app:solver_selection}

To construct the progressive solver sets, we test six candidate solvers with different model sizes and generation costs. For each solver and subgoal, eight proof attempts are generated and checked by the proof assistant. We define Close@8 as the fraction of unresolved subgoals for which at least one of the eight attempts is accepted. Generation cost is measured in GPU-seconds per attempt.

Table~\ref{tab:solver_screening} shows the test results. On every dataset where it was evaluated, Goedel-Prover-V2-32B achieves the highest Close@8 among the screened solvers. Goedel-Prover-V2-8B retains relatively high solving capability at roughly half of its generation cost on Lean Workbook. They are selected as the expensive and medium-cost solvers in the progressive settings.

The lowest-cost solver is selected based on computation saving rather than Close@8 alone. For each candidate, we combine it with Goedel-Prover-V2-8B and use the measured solver success rates to allocate attempts. We then measure the generation computation required to reach the coverage obtained
by Goedel-Prover-V2-8B alone. Table~\ref{tab:low_cost_selection} reports the relative saving. Kimina-Prover-Distill-8B reduces computation on all four datasets at the full Goedel-Prover-V2-8B coverage, whereas the other candidates do not give consistent reductions. Therefore, Kimina-Prover-Distill-8B is used as the third low-cost solver.

\subsection{Predictor Training Details}
\label{app:predictor_training}

The cross-solver predictors $\Theta_A$ and $\Theta_B$ use Goedel-Prover-V2-8B as the backbone. We
fine-tune the model with LoRA using rank 32, scaling factor 64, and dropout 0.05. LoRA is applied to the $q$, $k$, $v$, $o$, gate, up, and down projections. Training uses bfloat16, a learning rate of $10^{-5}$, batch size 16 with micro-batch size 4, three epochs, and a maximum sequence length of 1,536. Each model is trained on one H100 GPU.

The training set contains 3,488 Lean Workbook subgoals, with 870 additional subgoals reserved for evaluation using a problem-level split. The predictors are trained from the formal subgoal only, without previous proof attempts or proof-assistant feedback. For every training subgoal, the target is the least expensive available solver that closes the subgoal in the recorded eight-attempt proof pools. If no available solver closes it, the target is \texttt{none}. In the three-solver setting, the resulting training-label distribution is 30\% Kimina-Prover-Distill-8B, 12\% Goedel-Prover-V2-8B, 6\% Goedel-Prover-V2-32B, and 52\% \texttt{none}. $\Theta_A$ uses class-weighted cross-entropy training loss and $\Theta_B$ uses plain cross-entropy. All other model and optimization settings are identical. We train five random seeds for each objective and solver setting.

\section{Predictor Analysis}
\label{app:predictor}

\subsection{Success Prediction on the Evaluation Set}

We first examine whether the predicted success probabilities can distinguish subgoals that a solver can close from those it cannot, and whether the predicted probabilities are well calibrated.

The predictor outputs a distribution over the least expensive solver that can close the subgoal and the \texttt{none} class. We convert this distribution into a success probability for each solver using Eq.~\ref{eq:solverprobability}. For example, the success probability of Goedel-Prover-V2-8B includes the probability assigned to both Kimina-Prover-Distill-8B and Goedel-Prover-V2-8B, since either class indicates that Goedel-Prover-V2-8B is sufficient. The probability that at least one  solver can close the subgoal is $1-q_{\mathrm{none}}$.

Table~\ref{tab:predictor_eval} reports AUROC, Brier score, and expected calibration error (ECE) results on the 870 Lean Workbook evaluation subgoals. AUROC measures how well the predictor ranks solvable subgoals above unsolvable ones, with larger values indicating better
discrimination. Brier score measures the error of the predicted probabilities, and ECE measures their calibration. For both Brier score and ECE, smaller values are better.

\begin{table*}[t]
\centering
\caption{Success prediction on Lean Workbook evaluation set in the three-solver setting. Each entry reports AUROC / Brier score / ECE. Higher AUROC and lower Brier score and ECE are better.}
\label{tab:predictor_eval}
\scriptsize
\setlength{\tabcolsep}{5pt}
\begin{tabular}{lcccc}
\toprule
Objective
& Any solver
& Kimina-8B
& Goedel-8B
& Goedel-32B \\
\midrule
Class-weighted CE ($\Theta_A$)
& 0.782 / 0.215 / 0.155
& 0.813 / 0.161 / 0.042
& 0.798 / 0.188 / 0.074
& 0.778 / 0.221 / 0.169 \\
Plain CE ($\Theta_B$)
& 0.789 / 0.189 / 0.052
& 0.822 / 0.157 / 0.040
& 0.808 / 0.177 / 0.051
& 0.784 / 0.191 / 0.059 \\
\bottomrule
\end{tabular}
\end{table*}

The two objectives have similar AUROC across all four predictions, indicating similar ability to distinguish solvable from unsolvable subgoals. Their main difference is calibration. Plain cross-entropy gives lower Brier score and ECE in every column. For Goedel-Prover-V2-32B, for example, ECE decreases from 0.169 with class-weighted cross-entropy to 0.059 with plain cross-entropy. This better calibration is useful in the later stage of proving, where BudgetPV compares predicted success probabilities across many previously attempted subgoals. The complementary benefit of class weighting
for initial solver assignment is examined in the next ablation.

\subsection{Effect of Class Weighting on Solver Assignment}

The class-weighted and plain cross-entropy predictors make different initial solver assignments. We examine the 394 Lean Workbook evaluation subgoals that can be closed by at least one of the three solvers. An assignment is under-assigned when the selected solver cannot close the subgoal but a more
expensive solver can, and over-assigned when a more expensive solver is selected although a less expensive solver can close it.

\begin{table}[t]
\centering
\caption{Initial solver-assignment behavior on 394 subgoals that
are closable by at least one solver.}
\label{tab:assignment_behavior}
\small
\setlength{\tabcolsep}{5pt}
\begin{tabular}{lccc}
\toprule
Objective & First choice closes & Over-assigned & Under-assigned \\
\midrule
Class-weighted CE ($\Theta_A$)
& 337 / 394 & 102 & 52 \\
Plain CE ($\Theta_B$)
& 319 / 394 & 48 & 72 \\
\bottomrule
\end{tabular}
\end{table}

Class weighting reduces under-assignment from 72 to 52 and increases the number of subgoals closed by the first solver choice from 319 to 337. This is useful in the initial stage, where assigning a solver that is too weak causes an additional failed attempt before a stronger solver is used.
Plain cross-entropy makes fewer over-assignments and provides better calibrated probabilities, which is more useful later when BudgetPV compares previously attempted subgoals using their absolute success estimates.

\subsection{Success Prediction During Live Proving}

In live proving, BudgetPV continues to estimate success after a subgoal has received failed proof attempts. We evaluate whether the predictor can still distinguish successful from unsuccessful attempts as the number of previous attempts increases. Table~\ref{tab:predictor_live} reports AUROC on decisions collected during the live experiments, grouped by the number of previous attempts on the same subgoal.

\begin{table}[t]
\centering
\caption{Success prediction during live proving, grouped by the number of
previous attempts on the same subgoal. Higher AUROC is better.}
\label{tab:predictor_live}
\small
\setlength{\tabcolsep}{4pt}
\begin{tabular}{lrrr}
\toprule
Predictor & Previous attempts & Attempts & AUROC \\
\midrule
$\Theta_A$ & 0    & 12,830 & 0.820 \\
$\Theta_A$ & 1    & 7,886  & 0.817 \\
$\Theta_A$ & 2--3 & 2,561  & 0.817 \\
\midrule
$\Theta_B$ & 0    & 3,504  & 0.827 \\
$\Theta_B$ & 1    & 7,857  & 0.782 \\
$\Theta_B$ & 2--3 & 35,079 & 0.707 \\
\bottomrule
\end{tabular}
\end{table}

For $\Theta_A$, AUROC remains approximately 0.82 from the first attempt through two to three previous attempts. $\Theta_B$ also retains useful discrimination after previous failures, with AUROC of 0.782 after one previous attempt and 0.707 after two to three attempts. These results show that the success estimates remain informative as BudgetPV revisits previously attempted subgoals.

\end{document}